\documentclass[12pt]{iopart}
\usepackage{iopams}
\usepackage{amssymb,color}
\usepackage{graphicx,epsfig}	
\usepackage{subfigure}
\usepackage{epstopdf}
\usepackage{amsmath}
\usepackage{enumerate}
\usepackage{harvard}
\usepackage{ulem}

\def\xfull{\x_{\mbox{{\tiny full}}}}
\newtheorem{rem}{Remark}[section]

                    {$\blacksquare$\vspace*{7pt}} 

\def\R{{\mathbb R}}

\def\f{\frac}

\def\q{\quad}

\def\bi{{\mathbf i}}

\def\x{\boldsymbol{x}}
\def\y{{\boldsymbol y}}

\def\bi{\begin{itemize}} \def\ei{\end{itemize}}
\def\be{\begin{eqnarray*}}
\def\ee{\end{eqnarray*}}

\def\etal{{\it et al }}

\def\0{{\mathbf 0}}

\newcommand{\beq}{\begin{equation}}
\newcommand{\eeq}{\end{equation}}

\def\eref#1{(\ref{#1})}

\def\XXint#1#2#3{{\setbox0=\hbox{$#1{#2#3}{\int}$ }
\vcenter{\hbox{$#2#3$ }}\kern-.55\wd0}}

\begin{document}

\title[DL for undersampled MRI]{Deep learning for undersampled MRI reconstruction}

\author{Chang Min Hyun\dag, Hwa Pyung Kim\dag, Sung Min Lee\dag, Sungchul Lee\ddag \footnote[5]{To whom correspondence should be addressed (sungchul@yonsei.ac.kr)} and Jin Keun Seo\dag}
\address{\dag Department of Computational Science and Engineering, Yonsei University, Seoul, Korea}
\address{\ddag Department of Mathematics, Yonsei University, Seoul, Korea}

\begin{abstract}
	This paper presents a deep learning method for faster magnetic resonance imaging (MRI) by reducing $k$-space data with sub-Nyquist sampling strategies and provides a rationale for why the proposed approach works well. Uniform subsampling is used in the time-consuming phase-encoding direction to capture high-resolution image information, while permitting the image-folding problem dictated by the Poisson summation formula. To deal with the localization uncertainty due to image folding, a small number of low-frequency $k$-space data are added. Training the deep learning net involves input and output images that are pairs of the Fourier transforms of the subsampled and fully sampled $k$-space data. Our experiments show the remarkable performance of the proposed method; only 29$\%$ of the $k$-space data can generate images of high quality as effectively as standard MRI reconstruction with the fully sampled data.
\end{abstract}

\maketitle

\section{Introduction}

Magnetic resonance imaging (MRI) produces cross-sectional images with high spatial resolution
using strong nuclear magnetic resonances, gradient fields, and hydrogen atoms inside the human body\cite{Lauterbur1973,seobook2014}.
MRI does not use damaging ionizing radiation like X-rays, but the scan takes a long time \cite{Haacke1999,Sodickson1997} and involves confining the subject in an uncomfortable narrow bow.
Shortening the MRI scan time might help increase patient satisfaction, reduce motion artifacts from patient movement, and reduce the medical cost. The MRI scan time is roughly proportional to the number of time-consuming phase-encoding steps in $k$-space. Many efforts have been made to expedite MRI scans by skipping the phase-encoding lines in $k$-space while eliminating aliasing,
a serious consequence of the Nyquist criterion violation \cite{Nyquist1928} that is caused by skipping.
Compressed sensing MRI and Parallel MRI are some of the techniques used to deal with these aliasing artifacts.  Compressed sensing MRI uses prior information on MR images of the unmeasured $k$-space data to eliminate or reduce aliasing artifacts.
Parallel MRI installs multiple receiver coils and uses  space-dependent properties of receiver coils to reduce aliasing artifacts \cite{Sodickson1997,Pruessmann1999,Larkman2001}.
This paper focuses solely on single-channel MRI for simplicity; hence, parallel MRI is not discussed.

\begin{figure}[h]
	\centering
	\includegraphics[width=0.5\textwidth]{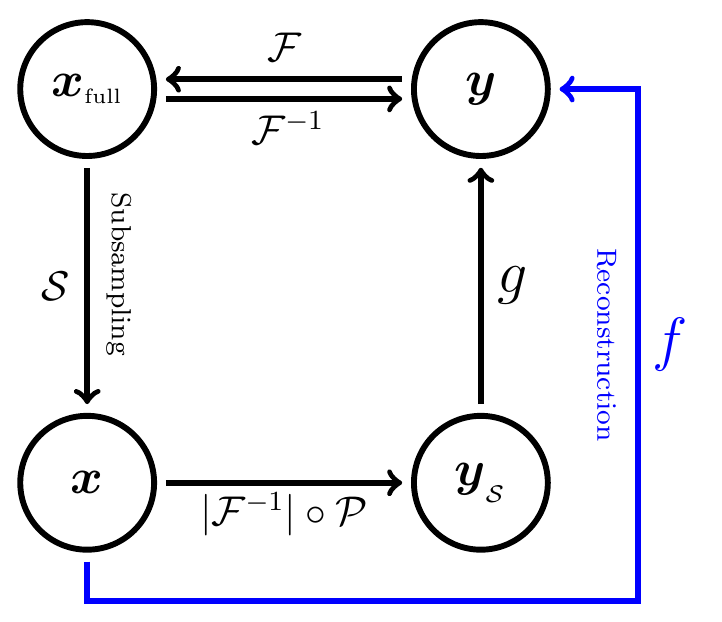}
	\caption{General strategy for undersampled MRI reconstruction problem. The inverse Fourier transform of a fully sampled $k$-space data $\xfull$ produces a reconstructed MRI image $\y$.  The goal is to find a subsampling function $\mathcal S$ and learn an undersampled MRI reconstruction $f$ from the training dataset. Here, $\y_{_{\mathcal S}}=|\mathcal F^{-1}|\circ\mathcal{P} (\x)$ is an aliased image caused by the violation of the Nyquist criterion. We use the U-net to find the function $g$ that provides the mapping from the aliased image $\y_{_{\mathcal S}}$ to an anti-aliased image $\y$.}\label{Fig-general-strategy}
\end{figure}

In undersampled MRI, we attempt to find an optimal reconstruction function $f: \x \mapsto \y$, which maps highly undersampled $k$-space data ($\x$) to an image ($\y$) close to the MR image corresponding to fully sampled data. Undersampled MRI consists of two parts, subsampling and reconstruction, as shown in Figure \ref{Fig-general-strategy}.

Compressive sensing (CS) MRI can be viewed as a sub-Nyquist sampling method in which the image sparsity is enforced to compensate for undersampled data \cite{Candes2006,Lustig2007}. CS-MRI can be described roughly as a model-fitting method to reconstruct the MR image $\y$ by adding a regularization term that enforces the sparsity-inducing prior on $\y$. It aims to reconstruct an image given by
\begin{equation}\label{leastSquare}
\y=\underset{\y}{\mbox{argmin}}\  \| \x - \mathcal S\circ\mathcal F(\y)\|_{\ell_2}^2 +\lambda \| \mathcal T(\y)\|_{\ell_1},
\end{equation}
where  $\mathcal F$ denotes the Fourier transform, $\mathcal S$ is a subsampling, $\mathcal T(\y)$ represents a transformation capturing the sparsity pattern of $\y$, $\circ$ is the symbol of composition,
and $\lambda$ is the regularization parameter controlling the trade-off between the residual norm and regularity. Here, the term $\| \x-\mathcal S\circ\mathcal F(\y)\|_{\ell_2}$ forces the residual $ \x-\mathcal S\circ\mathcal F(\y)$  to be small, whereas $ \| \mathcal T(\y)\|_{\ell_1}$ enforces the sparsity of $\mathcal T(\y)$.	
In CS-MRI, a priori knowledge of MR images is converted to a sparsity of  $\mathcal T(\y)$ with a suitable choice of $\mathcal T$.
The most widely used CS method is total variation denoising (i.e., $ \| \nabla \y\|_{\ell_1}$),
which enforces piecewise constant images by uniformly penalizing image gradients. Although CS-MRI with random sampling has attracted a large amount of attention over the past decade, it has some limitations in the preservation of fine-scale details and noise-like textures
that hold diagnostically important information in MR images.

In contrast to the regularized least-squares approaches \eref{leastSquare},
our deep learning approach is a completely reversed paradigm. It aims to learn a function $f:\x \mapsto \y$ using many training data $\{(\x^{(i)},\y^{(i)}):i=1,\cdots,N\}$. Roughly speaking, $f$ is achieved by
\begin{equation}\label{deep_eq}
f=\underset{f\in \Bbb U_{net}}{\mbox{argmin}}\   \frac1N \sum\limits_{i=1}^N \| f(\x^{(i)})-\y^{(i)}\|^2,
\end{equation}
where $\Bbb U_{net}$ is a deep convolutional neural network with some domain(or prior) knowledge determined by a training dataset that consists of pairs of fully sampled MR image and folded images. A U-net can provide a low-dimensional latent representation and preserve high-resolution features through concatenation in the upsampling process \cite{Ronneberger2015}.  This reconstruction function $f$ can be viewed as the inverse mapping of the forward model $\mathcal S \circ \mathcal F$ subject to the constraint of MR images, which are assumed to exist in a low dimensional manifold.
In the conventional regularized least-squares framework \eref{leastSquare}, it is very difficult to incorporate the very complicated MR image manifold into the regularization term.
However, in the deep learning framework,  the manifold constraint learned from the training set acts as highly nonlinear compressed sensing to obtain an useful reconstruction $f(\x)$ by leveraging complex prior knowledge on $\y$.

There are several recent machine learning based methods for undersampled MRI \cite{Ham2017arXiv,Kwon2017medphy,Ye2017arXiv} that were developed around the same time as our method. Hammernik \etal developed an efficient trainable formulation for an accelerated Parallel Imaging(PI)-based method of learning variational framework to reconstruct MR images from accelerated multicoil MR data. The method is designed to learn a complete reconstruction procedure for multichannel MR data in the regularized least-squares framework. Their aim is to learn a set of parameters associated with the gradient of the regularization in the gradient decent scheme. Kwon \etal applied the multilayer perceptron algorithm to reconstruct MR images from subsampled multicoil data. They reconstruct the image by using information from multiple receiver coils with different spatial sensitivities. In their method, the acceleration factor cannot be larger than the number of coils. Finally, Lee \etal used a residual learning method to estimate aliasing artifacts from distorted images of undersampled data.

In this paper, a subsampling strategy for deep learning is explained using a separability condition in order to produce MR images with a quality that is as high as regular MR image reconstructed from fully sampled $k$-space data. The subsampling strategy is to preserve the information in $\xfull$ as much as possible, while maximizing the skipping rate. To be precise, we use uniform subsampling in the phase encoding direction so that the Fourier transform contains all detailed features in a folded image, according to the Poisson summation formula.

We include a few low-frequency sampling to learn the overall structure of MR images and to deal with anomaly location uncertainty in the uniform sampling. The experiments show the high performance of the proposed method.

\section{Method} \label{sec_2}

Let $\y\in \Bbb C^{N\times N}$  be the MR image to be reconstructed, where $N^2$ is the number of pixels and  $\Bbb C$ is the set of complex numbers.
In 2D Fourier imaging with Cartesian  $k$-space sampling,
the MR image $\y$ can be reconstructed from the corresponding $k$-space data $\xfull\in \Bbb C^{N\times N}$: For $n,m=1-N/2, \cdots, 0, \cdots, N/2$,
\begin{equation}\label{MR-data-0}
\y(n,m)=\sum_{a=1-N/2}^{N/2}\sum_{b=1-N/2}^{N/2} \xfull(a,b) ~e^{2i\pi(an+bm)/N},
\end{equation}
where  $\xfull(a,b)$ is the MR-signal received at $k$-space position $(2 \pi a/N,2\pi b/N)$.
The frequency-encoding is along the $a$-axis
and the phase-encoding is along $b$-axis  in the $k$-space as per our convention.

In undersampled MRI,
we violate the Nyquist criterion and skip phase-encoding lines during the MRI acquisition to speed up the time-consuming phase encoding.
However, sub-Nyquist $k$-space data yields aliasing artifacts in the image space. For example, suppose we skip two phase-encoding lines to obtain an acceleration factor of 2. Then, the $k$-space data with zero padding is given by
{\begin{equation}\label{sampling}
	\left(
	\begin{array}{cccc}
	~\cdots~ &\xfull(\frac{N}{2}-1,\frac{N}{2}) & \xfull(\frac{N}{2},\frac{N}{2})
	\\ ~\cdots~ &0  & 0\\
	
	~\cdots~ &\xfull(\frac{N}{2}-1,\frac{N}{2}-2) & \xfull(\frac{N}{2},\frac{N}{2}-2)
	\\ ~\cdots~ &0  & 0\\
	\dots  &    \vdots &  \vdots\\
	\end{array}
	\right).
	\end{equation}}

According to the Poisson summation formula, the discrete Fourier transform of the above uniformly subsampled data with factor 2 produces the following two-folded image \cite{SeoWoo2012Wiley}:
\begin{equation}\label{Poisson}
\y_{2\mbox{\tiny -fold}}(n,m)=\y(n, m)+\y(n, m+N/2).
\end{equation}

If the deep learning approach is able to find an unfolding map $\y_{2\mbox{\tiny -fold}}\mapsto\y$, in this way
we could accelerate the data acquisition speed.
However, it is impossible to get this unfolding map even with sophisticated manifold learning for MR images.
In the left panel of Figure \ref{exphan},
we consider two different MR images $y_1$ and $ y_2$ with small anomalies at the bottom $(n,m)$ and top $(n, m+N/2)$, respectively. Here, the corresponding $k$-space data $\mathcal{F}(\y_1)$ and $\mathcal{F}(\y_2)$ are different. However, the corresponding uniformly subsampled $k$-space data with factor 2 $\mathcal{P} \circ \mathcal{S} \circ \mathcal{F} (\y_1)$ and $\mathcal{P} \circ \mathcal{S} \circ \mathcal{F}(\y_2)$ are completely identical because $\mathcal{F}^{-1} \circ \mathcal{P} \circ \mathcal{S} \circ \mathcal{F} (\y_1)= \mathcal{F}^{-1} \circ \mathcal{P} \circ \mathcal{S} \circ \mathcal{F} (\y_2)$. Here, $\cal S$ and $\cal P$ are the sampling and zero-padding operator, respectively, so that $\mathcal{P} \circ \mathcal{S} (\x_{full})$ is the subsampled $k$-space data with zero-padding given in \eref{sampling}.
It is not possible to identify whether the anomaly is at the top or bottom.
Deep learning cannot solve this unsolvable problem.
We now explain our undersampling strategy for deep learning.
\begin{rem}\label{Rem-undersample}
	Given the undersampled data $\x$, let $\y_\flat$ be the minimum norm solution, that is,
	$$
	\y_\flat=\underset{~~\y~\mbox{\footnotesize \rm s.t.~}  \mathcal{S} \circ \mathcal{F} \y = \x }{\mathrm{argmin}} \; \|\y\|_{\ell^2}.
	$$
	This $\y_\flat$ is $\mathcal{F}^{-1}(\mathcal{P}(\x))$, the inverse Fourier transform of the data $\x$ padded by zeros. This is because $\|\mathcal{P}(\x)\|_{\ell^2}\leq \|\x'\|_{\ell^2}$ for all $\x'$ satisfying $\mathcal{S} (\x')=\x$ and the Fourier transform map is an isometry with respect to the $\ell^2$ norm. Unfortunately, this minimum norm solution $\y_\flat$ is undesirable in most cases. See \ref{ap1}.
\end{rem}

\subsection{Subsampling Strategy} \label{sec_2.A}
\begin{figure}[h!]
	\centering
	\includegraphics[width=1\textwidth]{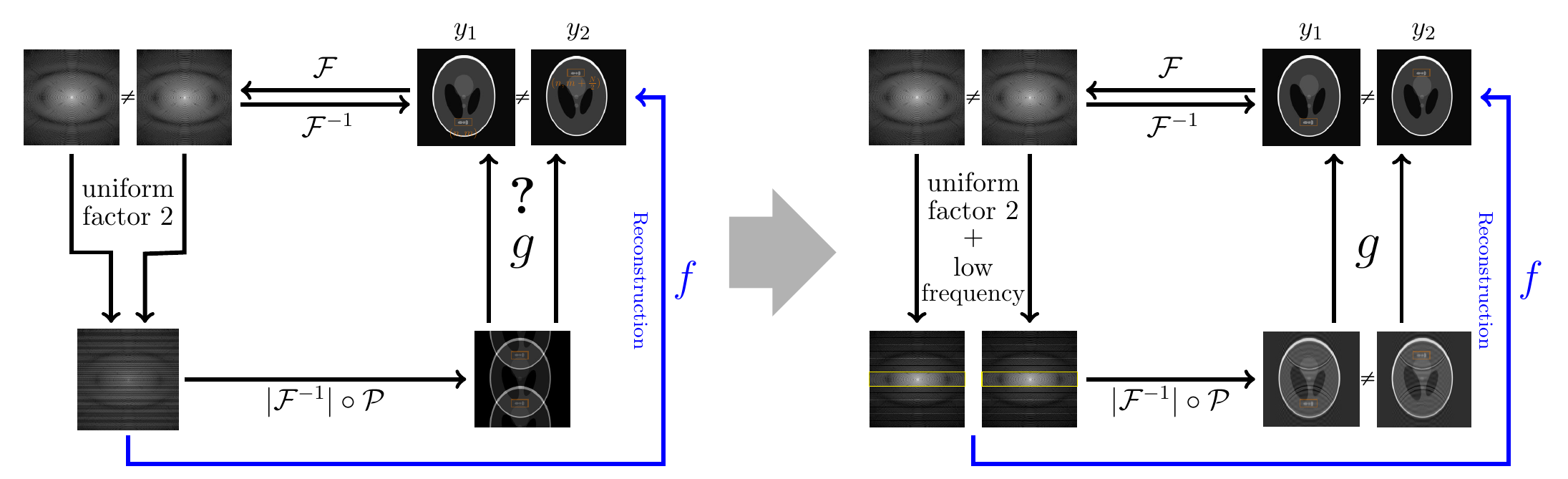}
	\caption{Feasibility of deep learning methods. Learning $f$ requires separability: $\y_1\neq \y_2  ~\mbox{implies }~ |\mathcal F^{-1}| \circ \mathcal P\circ \mathcal S \circ \mathcal F(\y_1)\neq |\mathcal F^{-1}| \circ \mathcal P\circ \mathcal S \circ \mathcal F(\y_2)$. The figure on the left shows why uniform subsampling does not satisfy the separability condition. We consider two different MR images with small anomalies at position $(n,m)$ and $(n, m+N/2)$, respectively. The corresponding $k$-space data are different, but the corresponding uniformly subsampled $k$-space data with factor 2 are completely identical. It is hence not possible to identify whether the anomaly is at the top or bottom. In contrast, the figure on the right shows why separability can be achieved by adding low frequency data. Additional low frequency lines in the yellow box provides the location information of small anomalies.}
	\label{exphan}
\end{figure}

Let $\{(\x^{(j)},\y^{(j)})\}_{j=1}^M$ be a training set of undersampled and ground-truth MR images.The vectors $\x^{(j)}$ and $\y^{(j)}$ are in the space $\Bbb C^{N\times N}$. Figure \ref{Fig-general-strategy} shows a schematic diagram of our undersampled reconstruction method, where the corresponding inverse problem is to solve the underdetermined linear system
\begin{equation}
\label{underdetermined} \; \mathcal S \circ \mathcal F (\y)~=~\x.
\end{equation}
Given undersampled data $\x$, there are infinitely many solutions $\y$ of \eref{underdetermined} in $\Bbb C^{N\times N}$.  It is impossible to invert the ill-conditioned system  $\mathcal S \circ \mathcal F: \Bbb C^{N\times N}\to \mathfrak{R}_{\mbox{\tiny$\mathcal S \circ \mathcal F$}}$, where $\mathfrak{R}_{\mbox{\tiny$\mathcal S \circ \mathcal F$}}$ is the range space of operator $\mathcal S \circ  \mathcal F$ and its dimension is much lower than $N^2$. We use the fact that the MR images of humans exist in a much lower-dimensional manifold $\mathcal M$ embedded in the space $\Bbb C^{N\times N}$. With this constraint $\mathcal M$ which is unknown, there is the possibility that there exists a practically meaningful inverse $f$ in the sense that
\begin{equation}
	\label{underdetermined2} \; f\left(\mathcal S \circ \mathcal F(\y)\right)=\y \q\mbox{for } \y\in \mathcal M.
\end{equation}

 In the left of Figure $\ref{exphan}$, we consider the case that $\mathcal S$ is the uniform subsampling of factor 2. With this choice of $\mathcal S$, two different images $\y_1\neq\y_2$ produce identical $ |\mathcal F^{-1}| \circ \mathcal P \circ \mathcal S \circ \mathcal F(\y_1)= |\mathcal F^{-1}| \circ \mathcal P \circ \mathcal S \circ \mathcal F(\y_2)$. This means the uniform subsampling of factor 2 is inappropriate for learning $f$ satisfying \eref{underdetermined2}. Here, $\y_1$ is the standard Logan phantom image and $\y_2$ is a modified image of $\y_1$ obtained by moving three small anomalies to their symmetric positions with respect to the middle horizontal line.
In contrast,
if we add a few low frequencies to the uniform subsampling of factor 2, as shown in the image on the right of Figure $\ref{exphan}$, the situation is dramatically changed and separability \eref{separability} may be achieved.
\begin{equation}
	\label{separability} \y_1\neq \y_2  ~~\mbox{implies }~~ |\mathcal F^{-1}| \circ \mathcal P \circ \mathcal S \circ \mathcal F(\y_1)\neq |\mathcal F^{-1}| \circ \mathcal P \circ\mathcal S \circ \mathcal F(\y_2).
	\end{equation}

\begin{figure}[h]
	\centering
	\includegraphics[width=1\textwidth]{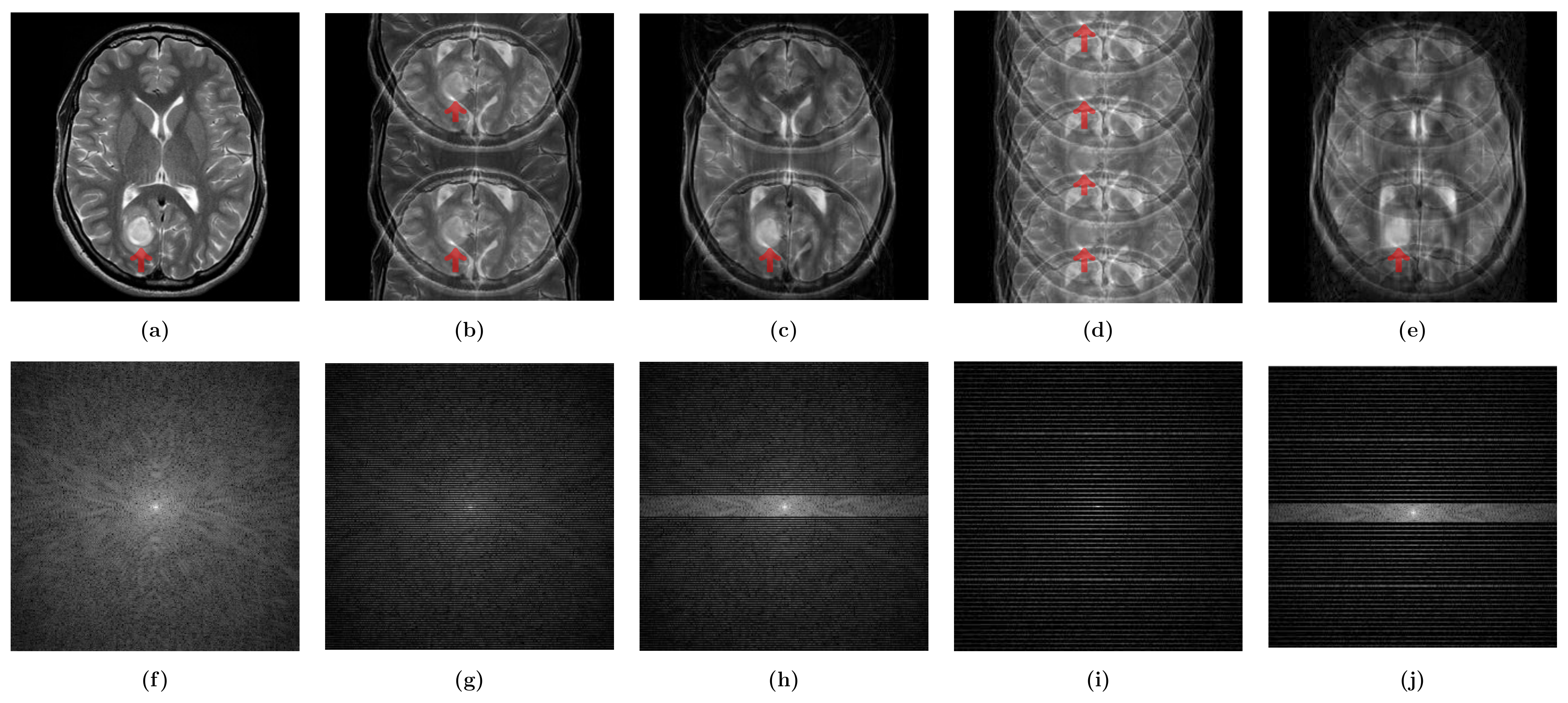}
	\caption{MR images of human brain with a tumor at the bottom. Images (a), (b), (c), (d), and (e) are reconstructed from (f) full sampling, (g) uniform subsampling of factor 2, (h) uniform subsampling of factor 2 with added some low frequencies,  (i) uniform subsampling of factor 4, and (j) uniform subsampling of factor 4 with added low frequencies , respectively. In (b) and (d), tumor-like lesions are found at both the top and bottom; one is a copy of the other. Hence, there exists a location uncertainty in the uniform sampling. However, in the reconstructed image (c) and (e) using the uniform subsampling of factor 2 and 4 with added low frequencies, the tumors are clearly located at the bottom. The location uncertainty can hence be addressed by adding a few low frequencies in $k$-space.}
	\label{lowfrequency}
\end{figure}

\begin{figure*}[h!]
	\centering
	\includegraphics[width=11cm,height=20cm]{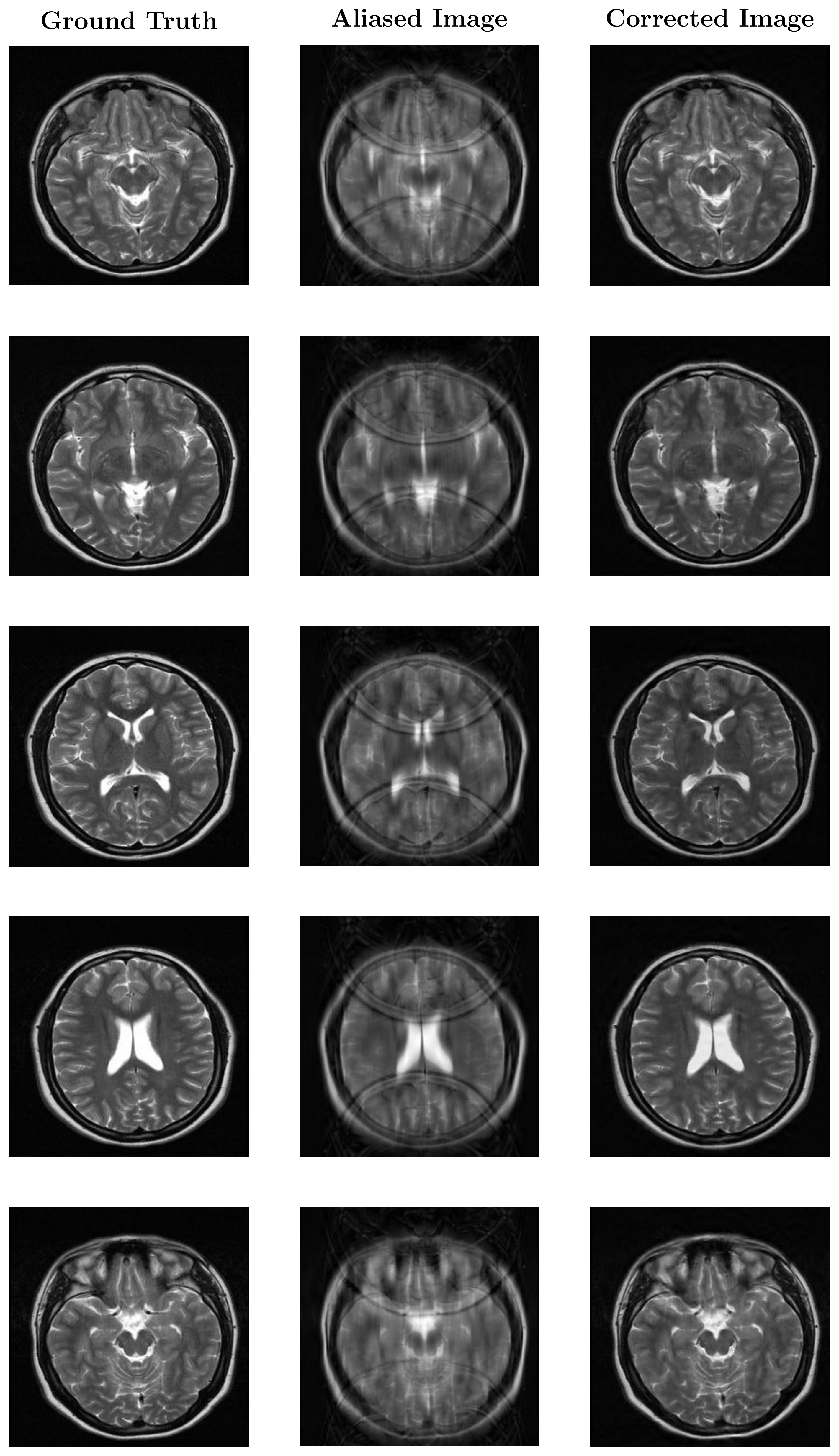}
	\caption{Numerical simulation results of five different brain MR images. The first, second and third columns show the ground-truth, aliased and corrected images, respectively. The proposed method significantly reduces the undersampling artifacts while preserving morphological information.}
	\label{result}
\end{figure*}

In Figure $\ref{lowfrequency}$, we demonstrate the separability condition again using the patient data.  Figure $\ref{lowfrequency}$ (a) is the ground truth, where the tumor is at the bottom. Figure $\ref{lowfrequency}$ (b) and Figure $\ref{lowfrequency}$ (d) are the reconstructed images using a uniform subsampling of factors 2 and 4, respectively; the tumors apear found at both the top and bottom, and the uniform subsampling of factor 2 and 4 are not separable. However, in the reconstructed images in Figure $\ref{lowfrequency}$ (c) and Figure $\ref{lowfrequency}$ (e) using the uniform subsampling of factosr 2 and 4 with added low frequencies, the tumors are clearly located at the bottom and separability \eref{separability} may be achieved. This crucial observation is validated by various numerical simulations as shown in Figure \ref{result}.
	
In the subsampling strategy, we use a uniform subsampling of factor 4 (25\% $k$-space data - 64 lines of a total 256 lines) with a few low frequencies(about $4 \% \; k$-space data - 12 lines of a total 256 lines).
Owing to the Poisson summation formula, the uniformly subsampled data with factor 4 provides the detailed structure of the folded image of $\y$ as
\begin{equation}\label{Poisson-4}
\y_{4\mbox{\tiny -fold}}(n,m)=\sum_{j=0}^3\y(n, m+\f{jN}{4}).
\end{equation}
However, the folded image may not contain the location information of small anomalies.
We fix the anomaly location uncertainty by adding a few amount of low frequency $k$-space data. (See appendix B for details.)

\subsection{Image Reconstruction Function}

\begin{figure*}[h!]
	\centering
	\includegraphics[width=1\textwidth]{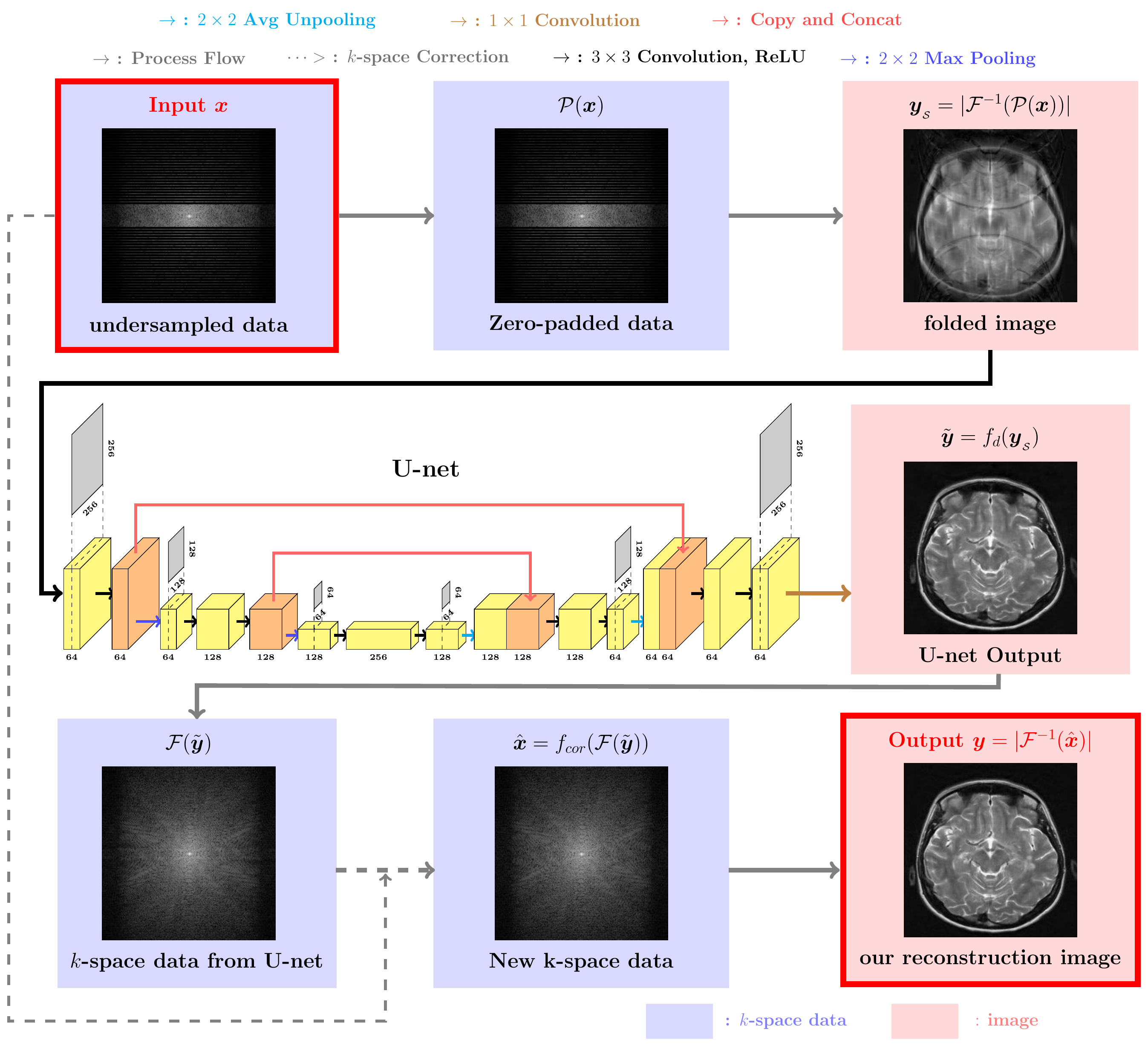}
	\caption{The proposed method consists of two major components : deep learning using U-net and $k$-space correction. As a preprecessing, we first fill in zeros for the unmeasured region of the undersampled data to get the zero-padded data. Then, we take the inverse Fourier transform, take its absolute value, and obtain the folded image. After the preprocess, we put this folded image into the trained U-net and produce the U-net output. The U-net recovers the zero-padded part of the $k$-space data. We take the Fourier transform and replace the unpadded parts by the original $k$-space data to preserve the original measured data. Finally, we obtain the final output image by applying the inverse Fourier transform and absolute value.}
	\label{U-net}
\end{figure*}

In this subsection, we describe the image reconstruction function $f$,
	which is schematically illustrated in Figure \ref{U-net}.
	When we have an undersampled data $\x$ as an input of $f$,
	about 70\% of $\x$ are not measured and not recorded.
	The first step of $f$ is to fill in zeros for the unmeasured region of $\x$ to obtain $\mathcal{P} (\x)$.
	After the zero padding has been added, we take the inverse Fourier transform of $\mathcal{P}(\x)$, take its absolute value, and obtain the folded image $\y_{_{\cal S}}$.
	We input this folded image $\y_{_{\cal S}}$ into the trained U-net and obtain the U-net output image $\tilde \y$.
	We apply the Fourier transform to $\tilde \y$, which yields  the $k$-space data ${\cal F}(\tilde \y)$.
	The U-net recovers the zero-padded part of the $k$-space information. However, during this recovery, the unpadded parts of the data are distorted. We manually fix this unwanted distortion by placing the original $\x$ values in their corresponding positions in the $k$-space data ${\cal F}(\tilde \y)$.
	We call this $k$-space correction as $f_{cor}$ and set $\hat\x=f_{cor}({\cal F}(\tilde \y))$. Because the original input data is preserved, we expect to obtain a more satisfactory reconstruction image and, indeed, our experiments show that the $k$-space correction is very effective. Finally, we apply the inverse Fourier transform to $\hat\x$, take the absolute value and obtain our reconstruction image $|{\cal F}^{-1}(\hat\x)|$.
	In summary, our image reconstruction function $f:\x\mapsto \y$ is given by
	\begin{equation}
	f= |\mathcal{F}^{-1}| \circ f_{cor} \circ \mathcal{F} \circ f_d \circ |\mathcal{F}^{-1}| \circ \mathcal{P},
	\end{equation}
	where
	$f_d$ is the trained U-net and $f_{cor}$ indicates the $k$-space correction. Here, $f_{d}$ should be determined by the following training process.

To train and test the U-net $f_{d}$, we generate the training and test sets as follows. Given ground-truth MR images $\{\y^{(j)}\}_{j=1}^N$, we take the Fourier transform of each $\y^{(j)}$, apply our subsampling strategy ${\cal S}$, which yields $\x^{(j)}$. This provides a dataset $\{(\x^{(j)},\y^{(j)})\}_{j=1}^N$ of subsampled $k$-space data and ground-truth MR images. The dataset is divided into two subsets : a training set $\{(\x^{(j)},\y^{(j)})\}_{j=1}^M$ and test set $\{(\x^{(j)},\y^{(j)})\}_{j=M+1}^N$. The input $\x^{(j)}$ of the image reconstruction function $f$ is an undersampled $k$-space data and the output $\y^{(j)}$ is the ground truth image. Using the zero-padding operator, inverse Fourier transform, and absolute value, we obtain folded images $\y_{_{\cal S}}^{(j)}$. Our training goal is then to recover the ground-truth images $\y^{(j)}$ from the folded images $\y_{_{\cal S}}^{(j)}$. Note that $\lbrace \y_{\cal S}^{(j)}, \y^{(j)} \rbrace_{j=1}^{M}$ is a set of pairs for training $f_d$.

The architecture of our U-net is illustrated in Figure \ref{U-net}. The first half of the network is the contracting path and the last half is the expansive path. The size of the input and output images is 256$\times$256. In the contracting path, we first apply the 3$\times$3 convolutions with zero-padding so that the image size does not decrease after convolution. The convolution layers improve the performance of machine learning systems by extracting useful features, sharing parameters, and introducing sparse interactions and equivariant representations \cite{Bengio2015book}. After each convolution, we use a rectified linear unit(ReLU) as an activation function to solve the vanishing gradient problem \cite{Glorot2011IEEE}. Then, we apply the 2$\times$2 max pooling with a stride of 2. The max pooling helps to make the representation approximately invariant to small translations of the input \cite{Bengio2015book}. In the expansive path, we use the average unpooling instead of max-pooling to restore the size of the output. In order to localize more precisely, the upsampled output is concatenated with the correspondingly feature from the contracting path. At the last layer a 1$\times$1 convolution is used to combine each the 64 features into one large feature \cite{Ronneberger2015}.

The input of the net is $\y_{_{\cal S}}^{(j)}$, the weights are $W$, the net, as a function of weights $W$, is $f_{net}(\cdot,W)$, and the output is denoted as $f_{net}(\y_{_{\cal S}}^{(j)},W)$. To train the net, we use the $\ell^2$ loss and find the optimal weight set $W_0$ with
\begin{equation}
 W_0=\underset{W}{\mbox{argmin}}\   \frac1M \sum\limits_{j=1}^M \| f_{net}(\y_{_{\cal S}}^{(j)},W)-\y^{(j)}\|_{\ell^2}^2.
\end{equation}
Once the optimal weight $W_0$ is found, we stop the training and denote the trained U-net as $f_d=f_{net}(\cdot,W_0)$.

In our experiment, the ground-truth MR image $\y$ was normalized to be in the range $[0,1]$ and the undersampled data $\x$ was subsampled to 29\% $k$-space data as described in Section \ref{sec_2}. We trained our model using a training set of 1,400 images from 30 patients. The MR images were obtained using a T2-weighted turbo spin-echo pulse sequence (repetition time = 4408 ms, echo time = 100 ms, echo spacing = 10.8 ms) \cite{Loizou2011}. To train our deep neural network, all weights were initialized by a zero-centered normal distribution with standard deviation 0.01 without a bias term.
The loss function was minimized using the RMSPropOptimize
with
learning rate 0.001,
weight decay 0.9,
mini-batch size 32, and
2,000 epochs.
RMSProp, which is an adaptive gradient method, was proposed by Tieleman and Hinton to overcome difficulties in the optimization process in practical machine learning implementations \cite{Hinton2012}. Training was implemented using TensorFlow \cite{Google} on an Intel(R) Core(TM) i7-6850K, 3.60GHz CPU and four NVIDIA GTX-1080, 8GB GPU system. The network required approximately six hours for training.

\section{Result} \label{Experiment}

Figure \ref{result} shows the performance of the proposed method for five different brain images in the test set. The first, second and third columns show the ground-truth, aliased and corrected images, respectively. The aliased images are folded four times. The proposed method suppresses these artifacts, but provides surprisingly sharp and natural-looking images.

\begin{figure}[h!]
	\centering
	\includegraphics[width=1\textwidth]{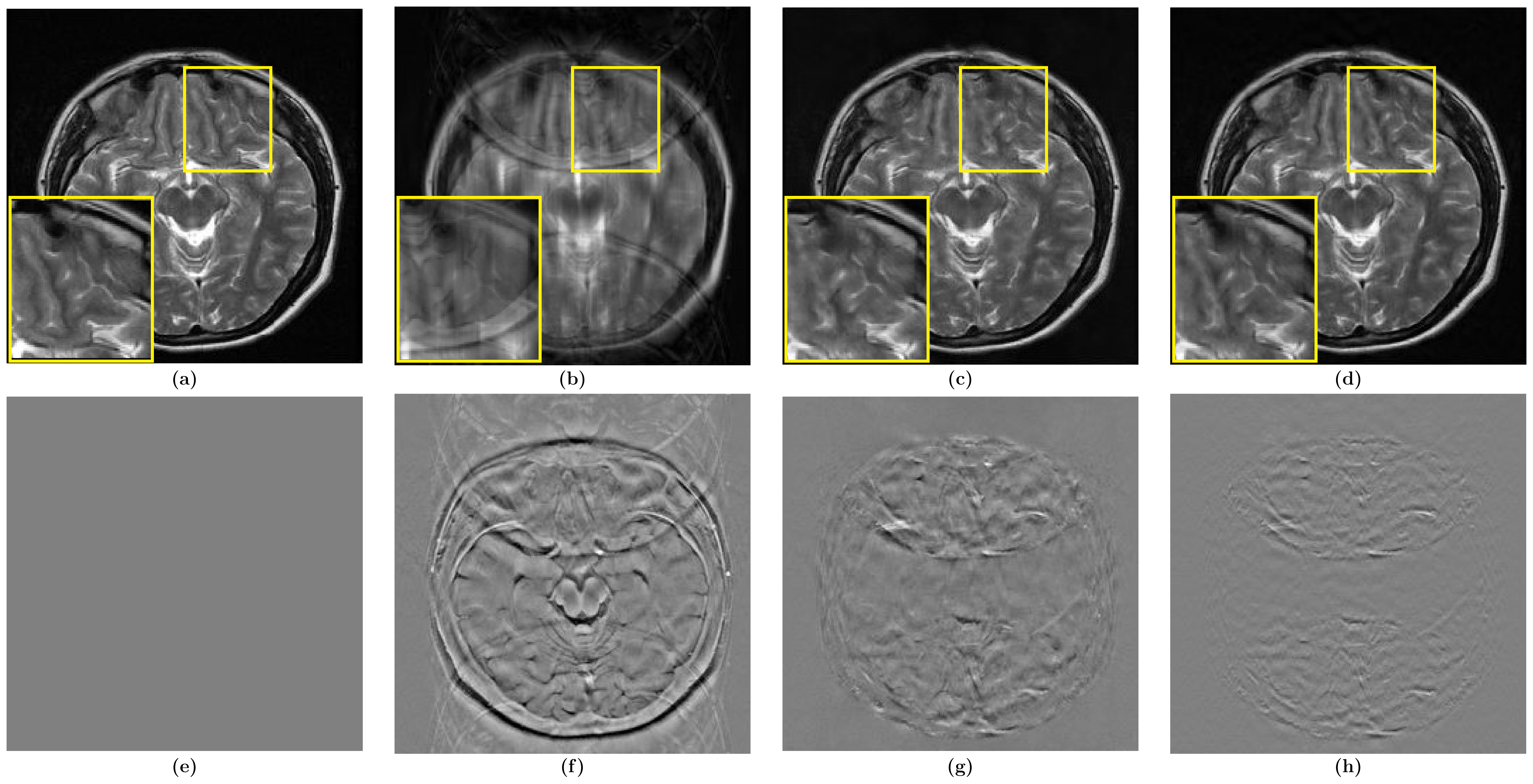}
	\caption{Simulation result using the proposed method : (a) ground-truth image, (b) aliased image, (c) output from the trained network, (d) $k$-space corrected image, Figure (e)--(h) depict the difference image with respect to the image in (a).}
	\label{reanddis}
\end{figure}

Figure \ref{reanddis} displays the impact of $k$-space correction.
The four images in the first row are the ground truth (Figure \ref{reanddis}(a)), input (Figure \ref{reanddis}(b)) and output (Figure \ref{reanddis}(c)) of the U-net, and the final output after the $k$-space correction (Figure \ref{reanddis}(d)). In the second row, we subtract  the ground truth from images in the first row.  Images Figure \ref{reanddis}(c) before and Figure \ref{reanddis}(d) after $k$-space correction are visually indistinguishable. However, Figures \ref{reanddis}(g) and (h) displays the impact of $k$-space correction. The U-net almost completely removes the folding artifacts. However, one can still see a few folding artifacts. Hence, The $k$-space correction removes the remaining folding artifacts.

 \begin{table}[h!]
 	\begin{center}
 		\caption{Quantitative evaluation results in terms of MSE and SSIM using the test set of 400 images. MSE is computed using  $\frac{1}{400\times 256^2} \sum_{i=1}^{400}\sum_{n=1}^{256}\sum_{m=1}^{256} (\y^{(i)}_{proposed}(n,m)-\y^{(i)}(n,m))^2$, where $\y^{(i)}$ is normalized to the range $[0,1]$. See \cite{SSIM2004} for definition of SSIM. As MSE approaches 0 or SSIM approaches 1, outputs are closer to labels.}
 		\label{result_tabel}
 		\begin{tabular}{c|c c c}
 			\hline
 			& Aliased  & U-net  & U-net with $k$-space correction  \\ \hline
 			MSE &  $0.0043 \pm 0.0016$ & $0.0012 \pm 0.0006$ & $0.0004 \pm 0.0002$ \\ 
 			SSIM &  $0.6516 \pm 0.0815$ & $0.8782 \pm 0.0411$ & $0.9039 \pm 0.0431$ \\ \hline
 		\end{tabular}
 	\end{center}
 \end{table}

All our qualitative observations are supported by the quantitative evaluation.
After we trained our model by using 1,400 images from 30 patients, we used a test set of 400 images from 8 other patients, and measure and report
their mean-squared error (MSE) and structural similarity index (SSIM) in Table \ref{result_tabel}.

The results for these metrics support the effectiveness of both the U-net and $k$-space correction. In particular, the effectiveness of $k$-space correction is demonstrated.

\section{Discussion and Conclusion}

Deep learning techniques exhibit surprisingly good performances in various challenging fields, and our case is not an exception. In this study, it generates the reconstruction function $f$ using the U-net, providing a better performance than the existing methods.

Our inverse problem of undersampled MRI reconstruction is ill-posed in the sense that there are fewer equations than unknowns. The underdetermined system in Section \ref{Experiment} has $256\times 256$ unknowns and $76\times 256$ equations. The dimension of  the set $\{\y\in \R^{256\times 256}: \mathcal S \circ \mathcal F (\y)= {\bf 0}\}$ is $(256-76)\times 256$, and therefore it is impossible to have an explicit reconstruction formula for solving (\ref{underdetermined}), without imposing the strong constraint of a solution manifold. For the uniqueness, the Hausdorff dimension of the solution manifold must be less than the number of equations (i.e., $76\times 256$). Unfortunately, it is extremely hard to find a mathematical expression for the complex structure of MR images in terms of $76\times 256$ parameters, because of its highly nonlinearity characteristic. The deep learning approach is a feasible way to capture MRI image structure as dimensionality reduction.

We learned the kind of subsampling strategy necessary to perform an optimal image reconstruction function after extensive effort. Initially, we used a regular subsampling with factor 4, but realized that it could not satisfy the separability condition. Because of wrap around artifact (a portion of the image is folded over onto some other portion of the image), it is impossible to  specify the locations of small objects. We added low frequencies hoping to satisfy separability and this turned out to guarantee separability in a practical sense.

Once the data set satisfies the separability condition, we have many deep learning tools to recover the images from the folded images. We chose to use the U-net.
The optimal choices may depend on the input image size, the number of training data, computer capacity, etc. It seems that the determination of optimal choice is difficult. Therefore, we empirically choose the number of layers, the number of convolution filters, and the filters' size.
The trained U-net successfully unfolded and recovered the images from the folded images.
The U-net removes most of the folding artifacts; however, one can still see them.
Hence, The $k$-space correction is used to further reduce them.

The experiments show that our learned function $f$ appears to have
highly expressive representation capturing anatomical geometry as well as small anomalies. We tested the flexibility of the proposed method. We applied the proposed method to CT images that were never trained. It worked well for different types of images that were never trained.
Our future research direction is to provide a more rigorous and detailed theoretical analysis to understanding why our method performs well.
The proposed method can be extended to multi-channel complex data for parallel imaging, with suitable modifications to the sampling pattern and learning network. This is our ongoing research topic. In practice, owing to the large size of input data available for deep learning, we may face  ``out of memory" problem. Indeed, we experienced out of memory problem when using input images of size $512 \times 512$, with a four GPU (NVIDIA GTX-1080, 8GB) system. This memory limitation problem was the primary reason to use $256 \times 256$ images, which were obtained by resizing $512 \times 512$ images. It is possible to develop more efficient and effective learning procedures for out of memory problem.

\section*{Acknowledgment}
This research was supported by the National Research Foundation of Korea No. NRF-2017R1A2B20005661. Hyun, Lee and Seo were supported by Samsung Science $\&$ Technology Foundation (No. SSTF-BA1402-01).

\newpage
\appendix
\section{Minimum-norm solution of the underdetermined system}\label{ap1}
The minimum-norm solution of the underdetermined system $\mathcal S\circ \mathcal F \y=\x$ in Remark 2.1 is the solution of following optimization problem: Minimize $\|y\|_{\ell^2}$ subject to the constraint $\mathcal S\circ \mathcal F \y=\x$. This underdetermined system has infinitely many solutions. For example, the following images are solutions of $\mathcal S\circ \mathcal F \y=\x$ where $\x$ is an undersampled data with a reduction factor of 3.37.
\includegraphics[width=1\textwidth]{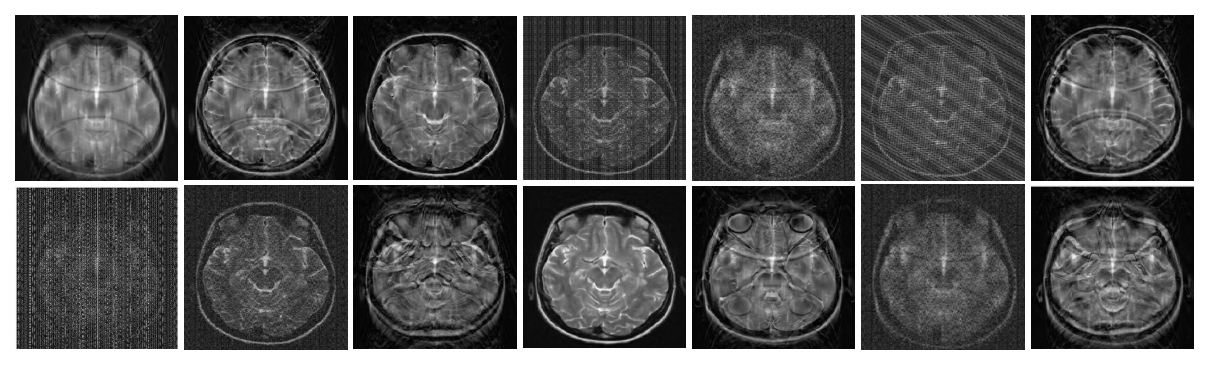}
The first image is the minimum-norm solution, i.e.,
\vskip -0.4in$$
\begin{array}{c}
\mbox{\tiny ~} \\
\includegraphics[width=0.12\textwidth]{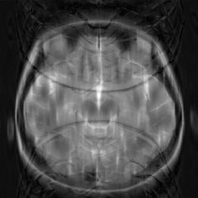}
\end{array}
=\mathcal{F}^{-1}(\x)=\underset{~~\y~\mbox{\footnotesize \rm s.t.~}  \mathcal{S} \circ \mathcal{F} \y = \x}{\mathrm{argmin}} \; \|\y\|_{\ell^2}
$$
This minimum-norm solution is improperly chosen; it does not look like a head MRI images. Then, can we deal with the complicated constraint problem: Solve $\mathcal S\circ \mathcal F \y=\x$ subject to the constraint that $\y$ looks like a head MRI image? It seems to be very difficult to express this constraint in classical logic formalisms.
\section{Performance of the proposed method with different reduction factors}
We tested the proposed method with different reduction factors from $R=3.37$ to $R=5.81$. We performed two experiments by varying two factors $\rho$ and $L$, where $\rho$ denotes the uniform subsampling rate along the phase encoding direction (vertical direction) and $L$ denotes the number of low frequency phase encoding lines to be added in our subsampling strategy.

\begin{figure}[h]
\centering
\includegraphics[width=0.85\textwidth]{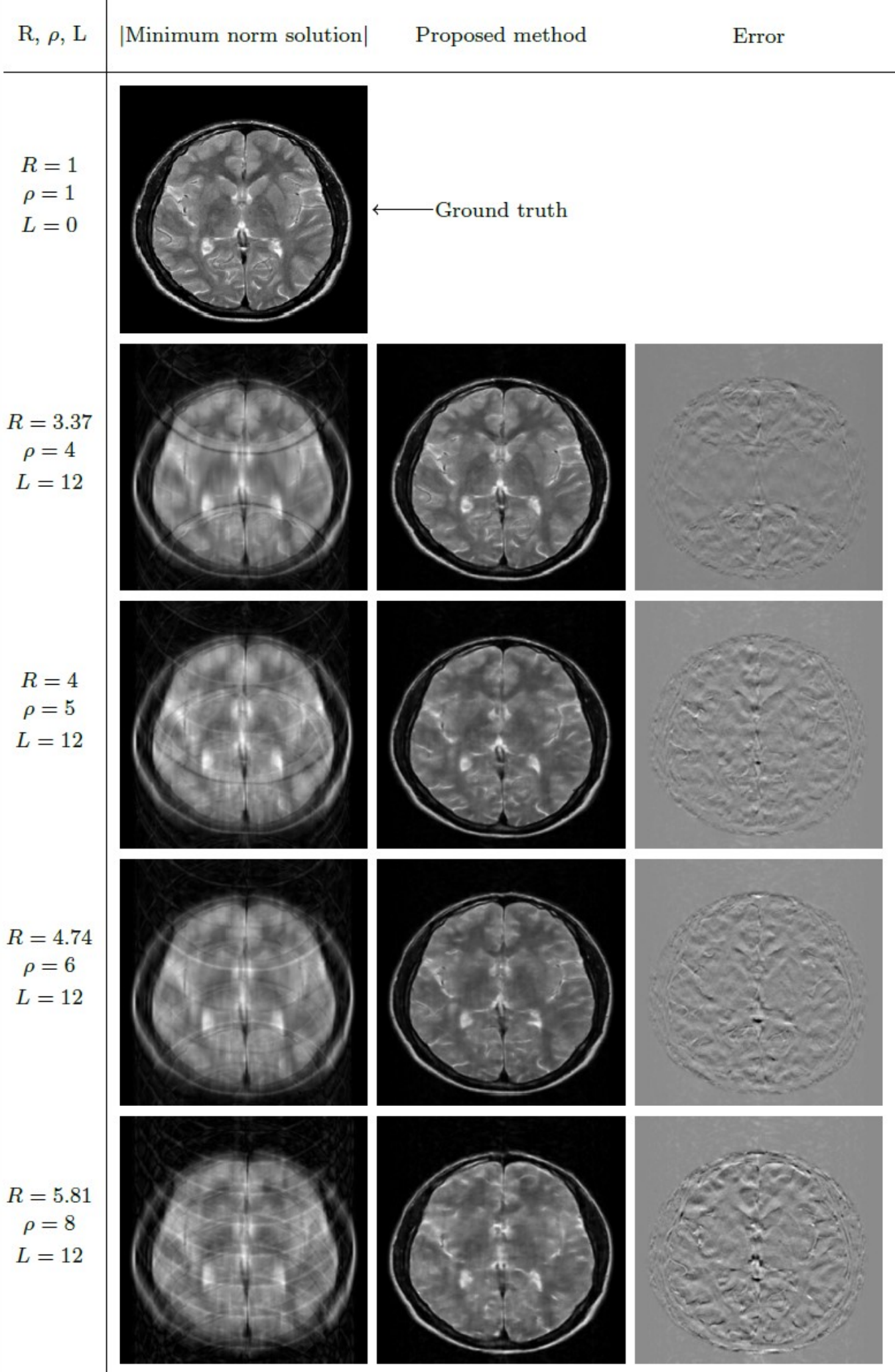}
\caption{In this experiment, we fix $L=12$ and vary $\rho$ : $\rho = 1, 4, 5, 6, 8$.}
\label{ex1}
\end{figure}

In Figure \ref{ex1}, we fix $L=12$ and vary $\rho$ from $\rho=4$ to $\rho=8$. The proposed method provides the good reconstruction image, even if $\rho$ is large $(\rho=8)$. See the last row in Figure \ref{ex1}.

\begin{figure}[h]
\centering
\includegraphics[width=0.85\textwidth]{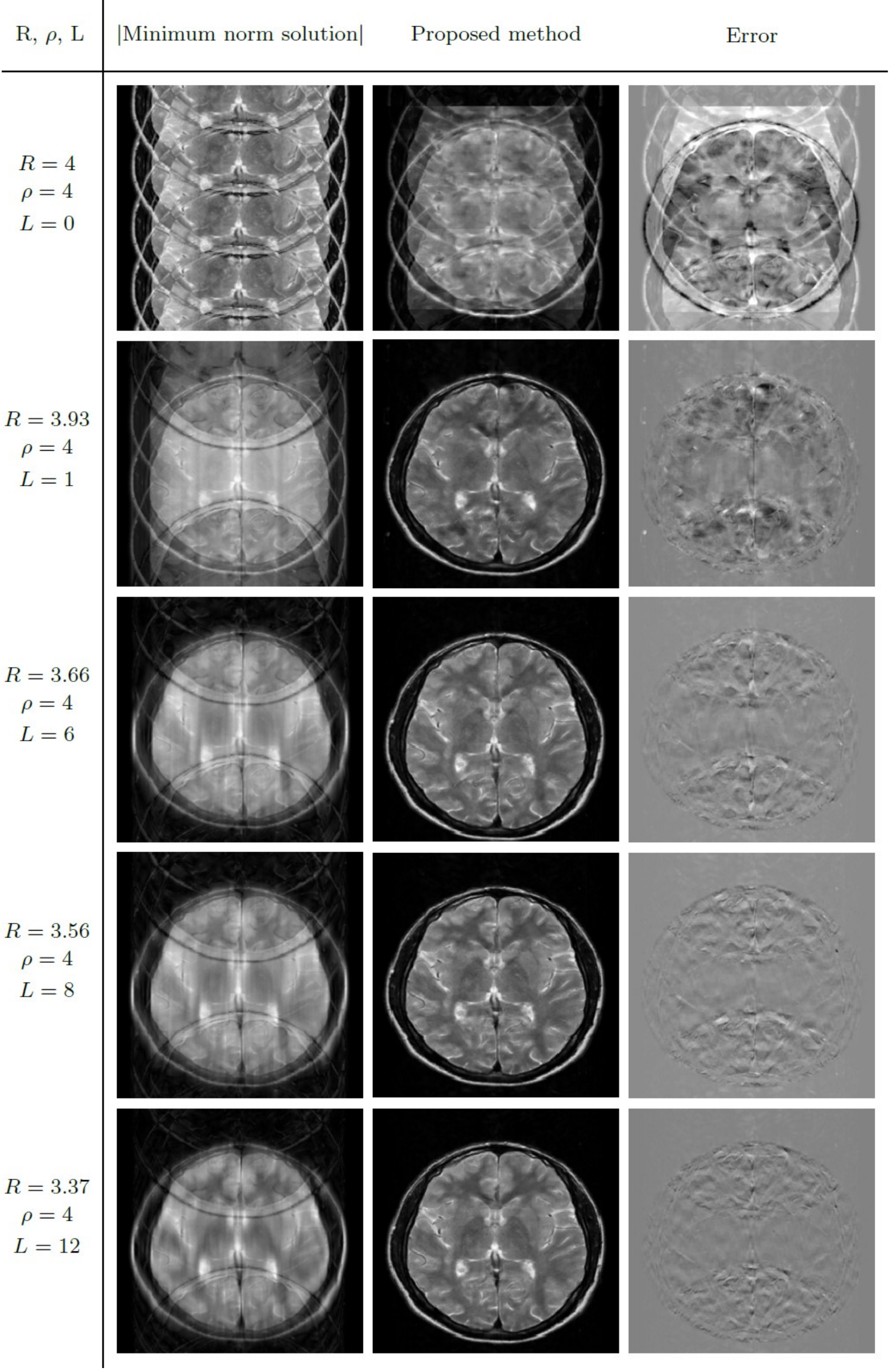}
\caption{In this experiment, we fix $\rho=4$ and vary $L$ : $L = 0, 1, 6, 8, 12$.}
\label{ex2}
\end{figure}

In Figure \ref{ex2}, we fix $\rho=4$ and vary $L$ from $L=0$ to $L=12$. In the case when the $L=0$, the separability condition is violated and the proposed method fails (as shown in the first row of Figure \ref{ex2}). When $L=1$, our network starts to learn unfolding, dramatically. The proposed method with $L=12$ provides excellent reconstruction capability.
\newpage
\newpage

\newpage
\section{The reconstruction process} This appendix presents the reconstruction process intuitively using a simplified version of the U-net.	
\begin{figure*}[h!]
\centering
\includegraphics[width=15cm, height=19cm]{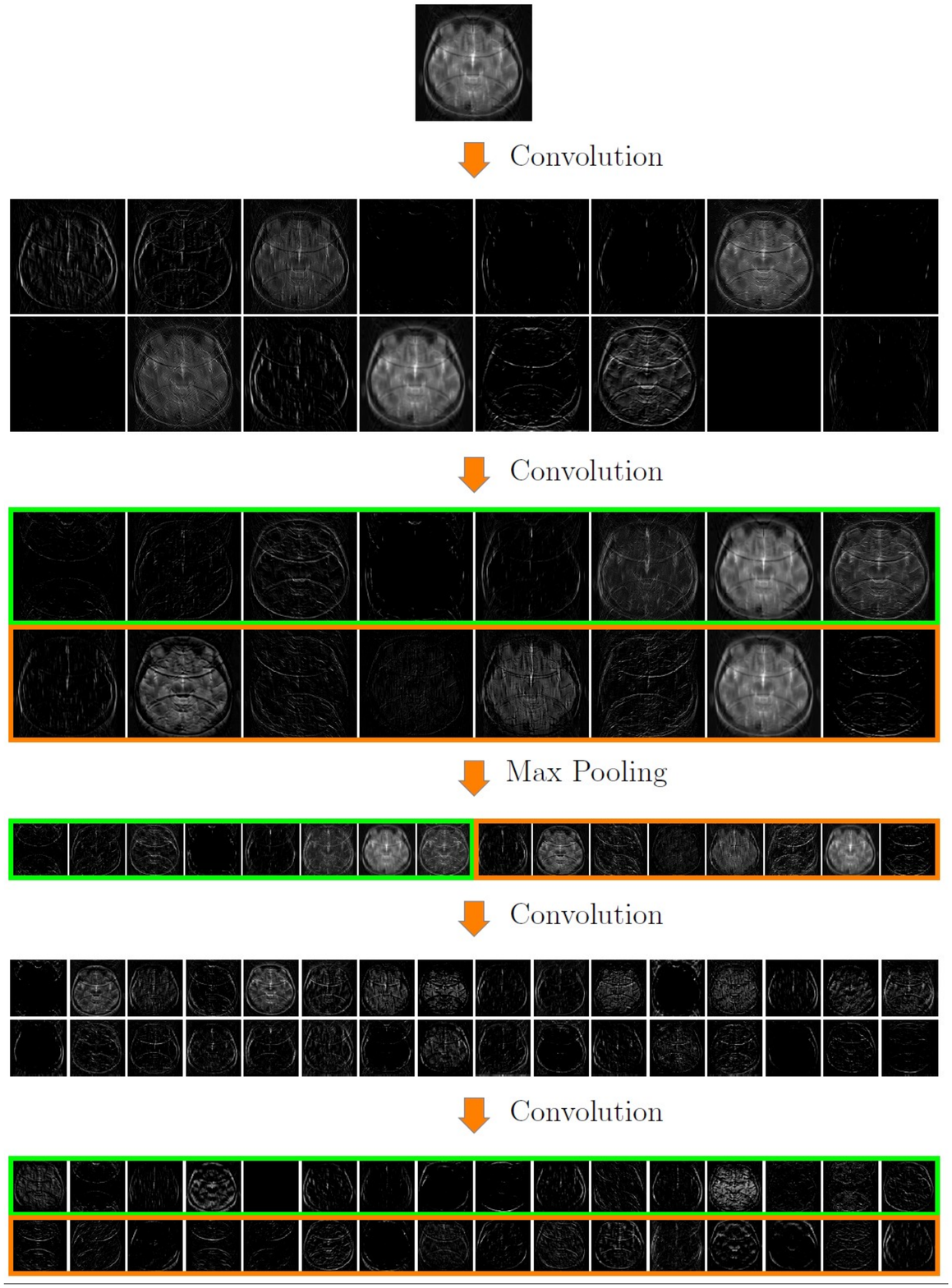}
\end{figure*}
	
\begin{figure*}[h!]
\centering
\includegraphics[width=15cm, height=22cm]{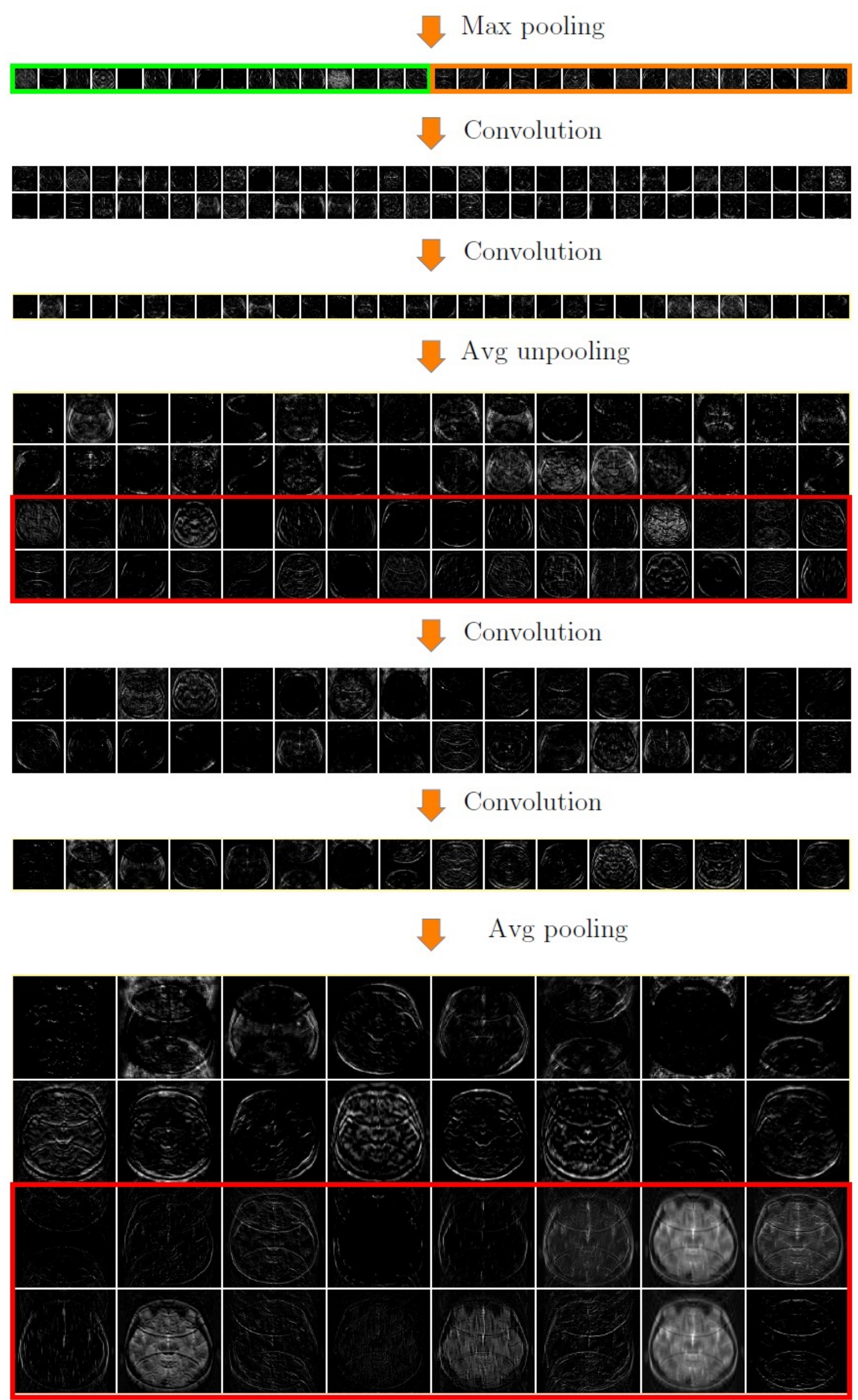}
\end{figure*}
	
\begin{figure*}[h!]
\centering
\includegraphics[width=15cm, height=22cm]{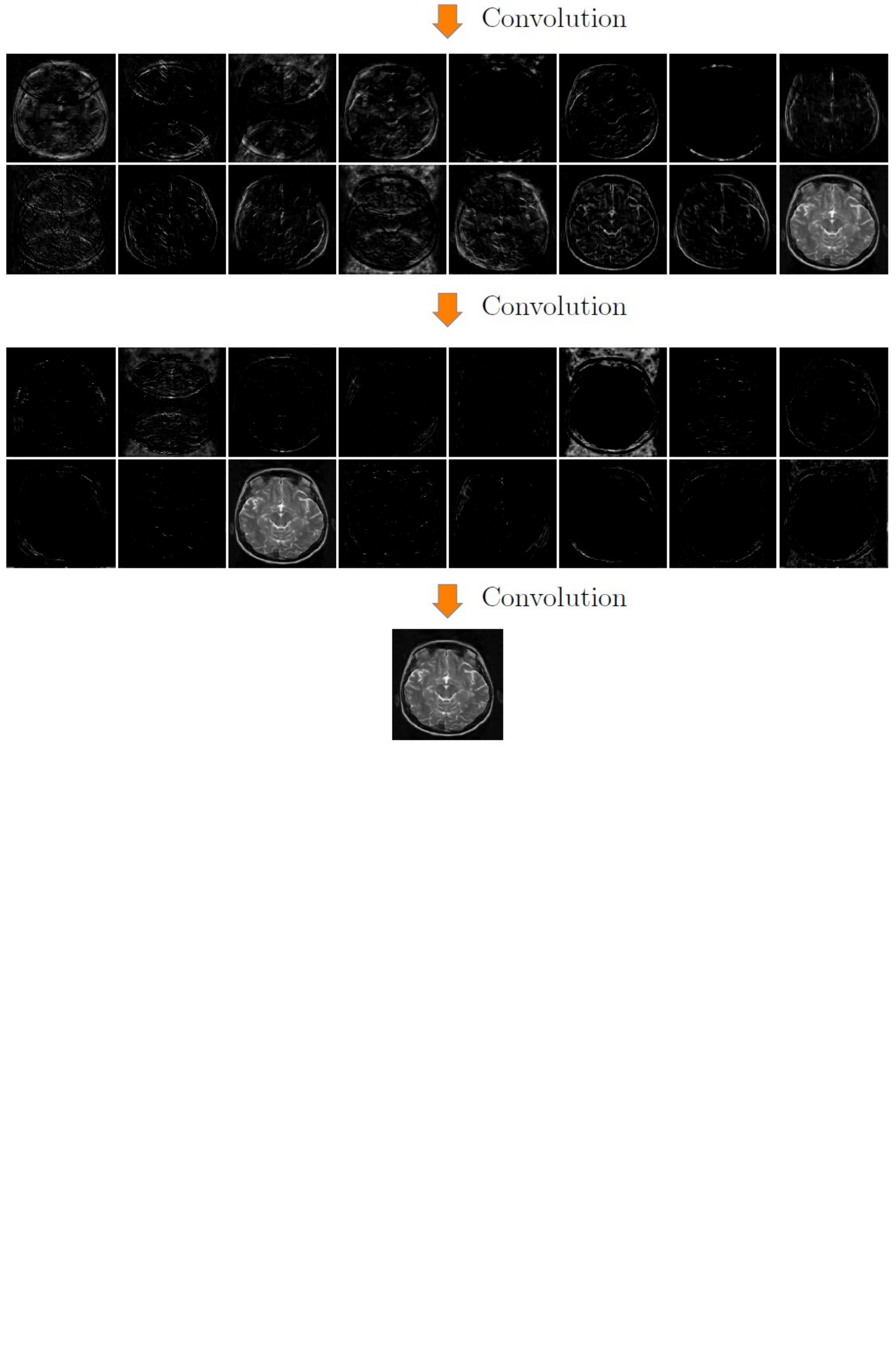}
\end{figure*}

\end{document}